\documentclass[conference]{IEEEtran}
\IEEEoverridecommandlockouts
\usepackage{cite}
\usepackage{amsmath,amssymb,amsfonts}
\usepackage{graphicx}
\usepackage{textcomp}
\usepackage[dvipsnames]{xcolor}
\usepackage{amsmath}
\usepackage{bbm}
\usepackage{algorithm}
\usepackage[noend]{algpseudocode}

\usepackage{subcaption}
\usepackage{balance}
\usepackage{comment}
\usepackage{enumerate}
\usepackage{amssymb,amsmath}
\usepackage{afterpage}
\usepackage{url}
\usepackage{xspace}
\usepackage{multirow}
\usepackage{multicol}
\usepackage{booktabs}
\usepackage{float}

\usepackage{enumitem}
\usepackage{mdwlist}
\usepackage{textcomp}
\usepackage{siunitx}

\allowdisplaybreaks

\newcommand{\figref}[1]{Fig.~\ref{#1}}
\newcommand{\figsubref}[1]{\subref{#1}}
\newcommand{\tabref}[1]{Table~\ref{#1}}

\newcommand{\secref}[1]{Sec.~\ref{#1}}

\renewcommand{\algref}[1]{Alg.~\ref{#1}}

\newcommand{\eg}{\emph{e.g.},\xspace}
\newcommand{\ie}{\emph{i.e.},\xspace}

\newcommand\fakeparagraph[1]{\par\noindent\textbf{{#1}}.\xspace}
\newcommand\fakeparagraphnodot[1]{\par\noindent\textbf{{#1}}\xspace}

\newcommand{\aucroc}{AUC-ROC\xspace}

\usepackage{color}

\def\BibTeX{{\rm B\kern-.05em{\sc i\kern-.025em b}\kern-.08em
    T\kern-.1667em\lower.7ex\hbox{E}\kern-.125emX}}
\begin{document}

\title{Class-dependent Pruning of Deep Neural Networks\\
}

\author{\IEEEauthorblockN{Rahim Entezari}
\IEEEauthorblockA{\textit{Institute of Technical Informatics, TU Graz } \\
\textit{CSH Vienna, Austria}\\
entezari@tugraz.at}
\and
\IEEEauthorblockN{Olga Saukh}
\IEEEauthorblockA{\textit{Institute of Technical Informatics, TU Graz } \\
\textit{CSH Vienna, Austria}\\
saukh@tugraz.at}
}

\maketitle

\begin{abstract}
Today's deep neural networks require substantial computation resources for their training, storage and inference, which limits their effective use on resource-constrained devices. Many recent research activities explore different options for compressing and optimizing deep models. On the one hand, in many real-world applications we face the \emph{data imbalance} challenge, \ie when the number of labeled instances of one class considerably outweighs the number of labeled instances of the other class. On the other hand, applications may pose a \emph{class imbalance} problem, \ie higher number of \emph{false positives} produced when training a model and optimizing its performance may be tolerable, yet the number of \emph{false negatives} must stay low. The problem originates from the fact that some classes are more important for the application than others, \eg detection problems in medical and surveillance domains.
Motivated by the success of the \emph{lottery ticket hypothesis}, in this paper we propose an iterative deep model compression technique, which keeps the number of false negatives of the compressed model close to the one of the original model at the price of increasing the number of false positives if necessary. Our experimental evaluation using two benchmark data sets shows that the resulting compressed sub-networks 1) achieve up to 35\% lower number of false negatives than the compressed model without class optimization, 2) provide an overall higher \aucroc measure compared to conventional Lottery Ticket algorithm and three recent popular pruning methods , and 3) use up to 99\% fewer parameters compared to the original network. The code is publicly available\footnote{\url{https://github.com/rahimentezari/Sensys-ml2020}}.
\end{abstract}


\section{Introduction}
\label{sec:intro}
While deep networks are a highly successful model class, their large memory footprint puts considerable strain on energy consumption, communication bandwidth and storage requirements of the underlying hardware, and hinders their usage on resource-con-\linebreak[0]{}strained IoT devices. For example, VGG-16 models for object detection~\cite{simonyan2014very} and facial attribute classification~\cite{lu2017fully} both contain over 130M parameters. 
Recent efforts on deep model compression for embedded devices explore several directions including quantization, factorization, pruning, knowledge distillation, and efficient network design. 
The approach presented in this paper combines network pruning with efficient network design by additionally including \emph{class-dependency} into the network compression algorithm. 

This is particularly useful in applications that can tolerate a slight increase in the number of \emph{false positives} (FP), yet need to keep the number of \emph{false negatives} (FN) close to the original model when pruning the weights.
Many real-world applications, \eg medical image classification and anomaly detection in production processes, have to deal with both \emph{data imbalance} and \emph{class imbalance} when training a deep model. On the one hand, real-world data often follows a long-tailed data distribution in which a few classes account for the most of the data, while many classes have considerably fewer samples~\cite{hasenfratz2014percom}. Models trained on these data sets are biased toward dominant classes~\cite{cui2019class}. Related literature treats data imbalance as a problem which leads to low model quality~\cite{chawla2002smote}. The proposed solutions typically adopt class re-balancing strategies such as re-sampling~\cite{buda2018systematic} and re-weighting~\cite{wang2017learning} based on the number of observations in each class. On the other hand, there are many TinyML applications, which have unbalanced class importance: \eg missing an event may have far more severe consequences than triggering a false alarm. This is especially the case in many detection scenarios and early warning systems in the IoT domain. Our method automatically shrinks a trained deep neural network for mobile devices integration. In this paper, we focus on \emph{keeping the number of FN low} when compressing a deep network.

To the best of our knowledge, this is the first paper proposing a class-dependent model compression. We provide an end-to-end network compression method based on the recently proposed iterative \emph{lottery ticket} (LT) algorithm~\cite{frankle2018lottery} for finding efficient smaller subnetworks in the original network. Since data imbalance is a common problem model designers have to deal with, \emph{in the first step}, we extend the original LT algorithm with a parameterized loss function to fight data imbalance. Related literature suggests that a direct compensation for data imbalance into the loss function outperforms alternative methods~\cite{cui2019class}. \emph{In the second step}, we control the number of false negatives and false positives of the model by including a parameterized \aucroc measure (see \secref{sec:exp}) into the model compression task. We observe that it is beneficial to train the first epoch for balanced classes to learn the class boundaries. In the later epochs, however we focus on class-imbalanced optimization with the goal to keep the number of FN at the level of the original model. We evaluate the new method on two public data sets and show that it achieves up to 35\,\% lower number of FN than the original LT algorithm without class-imbalanced training while preserving only 1\,\% of the weights. Surprisingly, our method with a novel loss function consistently outperforms the original version of the LT algorithm in all tested cases.

\section{Class-dependent Network Compression}
\label{sec:method}

\begin{figure}[t]
\includegraphics[width=\columnwidth]{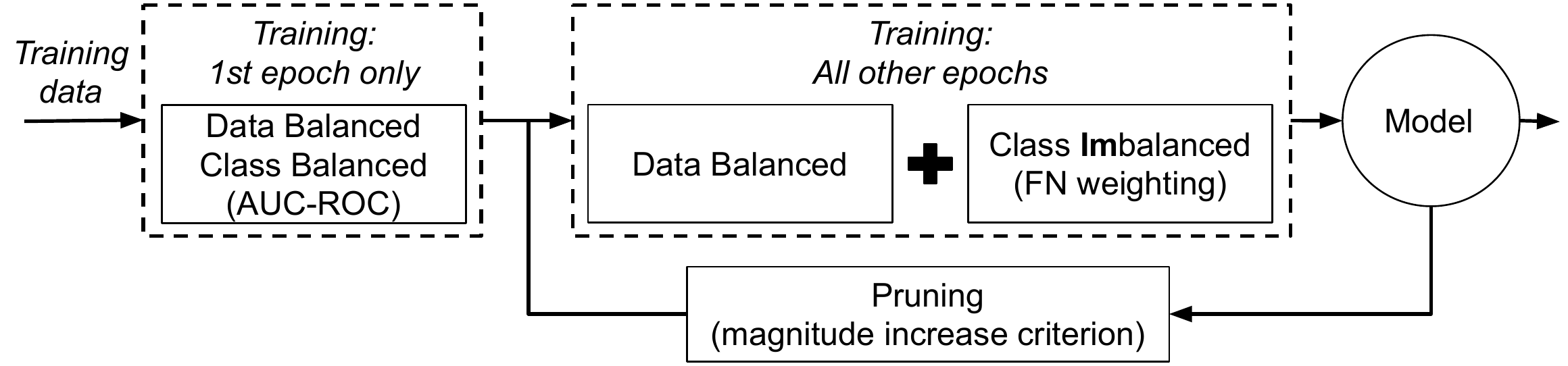}
\caption{Iterative network pruning with FN optimization.}
\label{fig:method}
\end{figure}

Our proposed method consists of the network compression pipeline presented in \figref{fig:method}. Training data is used to first train a class-balanced model, while further epochs prefer minimizing FN over FP. Our optimization function presents a combination of two loss functions in order to 1) control data imbalance, and 2) control class imbalance. We adopt a parameterized cross-entropy loss~\cite{cui2019class} to achieve the former, and use the hinge ranking loss~\cite{steck2007hinge} to maximize \aucroc and to control the trade-off between FN and FP to address the latter. The iterative pruning procedure is based on the LT algorithm. The paragraphs below describe the individual building blocks of our solution.

\fakeparagraph{Lottery Ticket (LT) algorithm}
Our class-dependent network compression method leverages the recently introduced iterative pruning used to search for efficient sparse sub-networks called \emph{lottery tickets} (LT)~\cite{frankle2018lottery} within an original deep network. LT networks often show superior performance when compared to the original network. The recently conducted analysis of LT networks~\cite{zhou2019deconstructing} suggests that the sub-network structure tightly coupled with network initialization is crucial to achieve high performance of LT networks. Moreover, the following conditions are responsible for the best results: 1) setting pruned values to zero rather than to any other value, 2) keeping the sign of the initialization weights when rewinding, and 3) keeping the weights with the largest absolute values or apply the magnitude increase mask criterion during iterative pruning. We leverage all these findings in this work. We use the magnitude increase criterion throughout the paper, \ie we rank the differences between the final and the initial values of the network weights in every round and prune the least \emph{p} percent.

\begin{algorithm}[h]
\caption{Class-dependent network compression}
\label{alg:LT}
\begin{algorithmic}[1]
\State Randomly initialize the network $f(x;m\bigodot W_{0})$ with the initially trivial pruning mask $m = 1^{|W_{0}|}$;
\State Train the network for $n$ iterations with the class-dependent loss $\mathcal{L}$ producing the network $f(x;m\bigodot W_{k})$;
\State Prune $p$\% of the remaining weights with the magnitude increase strategy, \ie $m[i] := 0$ if $W_{k}[i]$ gets pruned;
\State Replace remaining weights $W_k$ with their initial values $W_0$;
\State Go to step 2 if the next ($k+1$)-th round is required.
\end{algorithmic}
\end{algorithm}

\algref{alg:LT} provides a pseudo-code of the LT algorithm with the magnitude increase mask criterion and a class-dependent loss function $\mathcal{L}$ explained below. The algorithm initializes the network with random weights $W_0$ and applies an initially trivial pruning mask $m = 1^{|W_{0}|}$. The operation $\bigodot$ denotes an element-wise multiplication. After training the network for $n$ iterations we prune $p$ percent of the weights using the magnitude increase strategy by updating the mask $m$ accordingly. The remaining weights $W_k$ are then reset to their initial values $W_0$ before the next algorithm round starts.

In every round of the algorithm we minimize the loss function $\mathcal{L}$ of the following form
\begin{equation}
  \label{eq:loss}
	\mathcal{L} = \mathcal{L}_{wCE} + \mathcal{L}_{SHR},
\end{equation}
\noindent where $ \mathcal{L}_{wCE}$ and $ \mathcal{L}_{SHR}$ are the \emph{weighted binary cross-entropy loss} and the \emph{squared hinge ranking loss} respectively detailed below. 

\fakeparagraph{Weighted binary cross-entropy loss}
Inspired by~\cite{cui2019class}, we extend the notion of the classical binary cross-entropy to compensate for the data imbalance by introducing per-class weighting coefficients as follows
\begin{equation}
  \label{eq:BCE}
	\mathcal{L}_{wCE} = -\sum_{c=1}^M \gamma_c \cdot y_{o,c}\log(p_{o,c}),
\end{equation}
\noindent where $\gamma_c$ are weighting coefficients for every class; $M$ is the number of classes; $y_{o,c} \in \{0,1\}$ is a binary indicator if the class label $c$ is a correct classification of the observation $o$; $p_{o,c}$ is a predicted probability that the observation $o$ is of class $c$, and $n_c$ is the number of samples in $c$. 

We leverage the results in \cite{huang2016learning,wang2017learning} and handle the weighting coefficients $\gamma_c$ for individual classes as $\gamma_c = \frac{1-\beta}{1-\beta^{n_c}}$, where $\beta \in [0; 1)$ is a hyperparameter. In contrast to their work, we choose the value of the hyperparameter $\beta$ to compensate the data imbalance in favor of a particular class. The setting $\beta=0$ corresponds to no weighting and $\beta\rightarrow1$ corresponds to weighting by inverse data frequency for a given class. 
%
Recent work by \cite{cui2019class} shows that the weighting coefficients $\gamma_c$ play an important role in data-balancing. In particular, when training a CNN on imbalanced data, data-balancing for each class, by means of $\gamma_c$, provides a significant boost to the performance of the commonly used loss functions, including cross-entropy. 

\fakeparagraph{Squared hinge ranking loss}
The previous literature shows that 1) optimizing classification accuracy by minimizing cross-entropy cannot guarantee maximization of \aucroc~\cite{cortes2004auc}, and 2) \aucroc maximization as an optimization task yields a discontinuous non-convex objective function and thus cannot be approached by the gradient based methods \cite{yan2003optimizing}. Proposed solutions for \aucroc maximization~\cite{gultekin2018mba} are based on approximations. In this paper, we use the squared hinge ranking loss suggested in~\cite{steck2007hinge}, while adding the parameters $\lambda_c$ to control the class imbalance.


\begin{equation}
  \label{eq:squared_hinge_ranking_loss}
	\mathcal{L}_{SHR} = -\sum_{c=1}^M \lambda_c \max(1-y_{o,c}r_{o,c}, 0)^2.
\end{equation}

The squared hinge ranking loss is obtained from the hinge loss by replacing $p_{o,c}$ by a sorted in ascending order classifier output $r_{o,c}$. 

The authors of \cite{steck2007hinge} show that \aucroc can be written in terms of the hinge rank loss as follows 
\begin{equation}
	\text{\aucroc} \geq 1-\dfrac{{L}_{SHR}-C}{n^{+} n^{-}},
\end{equation}
where $n^+, n^-$ are the number of positive and negatives samples and $C$ is a constant independent of the rank order. Minimizing hinge ranking loss leads to \aucroc maximization ~\cite{steck2007hinge}.  
We use the \emph{squared} hinge ranking loss $\mathcal{L}_{SHR}$ to ensure our loss function $\mathcal{L}$ is differentiable.

In the next section we evaluate the effectiveness of the proposed class-dependent network compression algorithm on two benchmark data sets.

\section{Experimental Evaluation}
\label{sec:exp}

This section introduces the benchmark data sets, lists the metrics we use to evaluate the performance of our method, justifies parameter choices, and presents the results.

\subsection{Data Sets} 
We chose our benchmark data sets based on the complexity of the classification task they present.
ISIC is a challenging medical imaging lesion classification data set introduced in a data science competition\footnote{\url{http://challenge2016.isic-archive.com}}. The CRACK data set enjoys low classification complexity. Both benchmarks are introduced below.
\begin{figure}[t]
\includegraphics[width=\columnwidth]{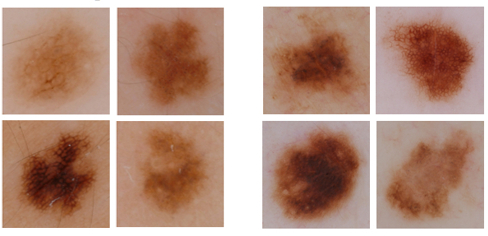}
\caption{Images from the ISIC data set~\cite{gutman2016skin}: benign (left) and melanoma (right) samples.}
\label{isic}
\end{figure}

\begin{figure}[t]
\includegraphics[width=\columnwidth]{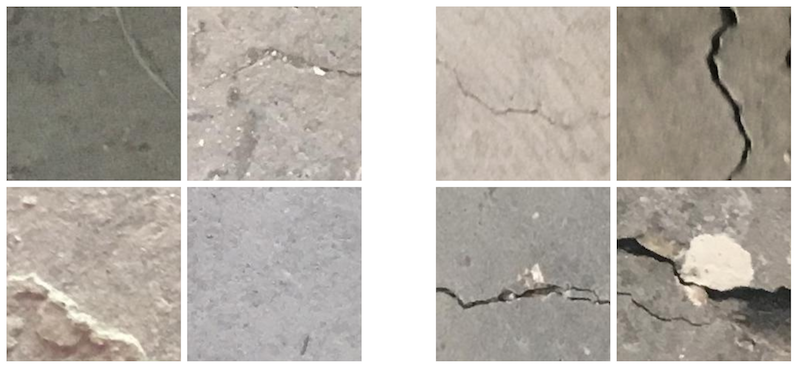}
\caption{Images from the CRACK data set~\cite{ozgenel2017concrete}: negative (left) and positive (right) samples.}
\label{crack_pos_neg}
\end{figure}

\fakeparagraphnodot{ISIC-2016} lesion classification dataset~\cite{gutman2016skin} includes original medical images paired with a confirmed malignancy diagnosis labels obtained from expert consensus and pathology report information, \ie each image is assigned a label either benign or melanoma. The training data set contains 900 dermoscopic lesion images with 173 positive and 727 negative examples respectively, whereas the test set includes 379 images with 76 positive and 303 negative samples respectively. \figref{isic} shows positive and negative sample images from the ISIC data set. 
\begin{table}[t]
\centering
\begin{tabular}{p{0.14\columnwidth}p{0.77\columnwidth}}
\hline
Name & Description \& Value\\
\hline 
Saturation & Modify saturation by 0.3 \\ 
Contrast & Modify contrast by 0.3 \\
Brightness & Modify brightness by 0.3 \\
Hue & Shift the hue by 0.1 \\
Flip & Randomly flip horizontally and vertically \\
Affine & Rotate by \ang{90}, shear by \ang{20}, scale by [0.8, 1.2] \\
Crop & Randomly crop ($>$40\% area) the image \\
Elastic & Warp images with thin plate splines \\
\hline
\end{tabular}
\caption{ISIC-2016 data augmentation~\cite{perez2018data}.} 
\label{table:augmentation}
\end{table}

We leverage best-practice data augmentation techniques suggested in~\cite{perez2018data}and shown in \tabref{table:augmentation}. This includes modification to image saturation, contrast, and brightness by 0.3, shifting the hue by 0.1, randomly flipping an image, Affine transformation (rotation by \ang{90}, shear by \ang{20}, scale by [0.8, 1.2]), randomly cropping an image ($>$40\% area). Saturation, contrast and brightness augmentations simulate changes in color due to camera settings and lesion characteristics. Affine transformations reproduce camera distortions and create new lesion shapes. 
Elastic warp is generated by defining the origins as an evenly-spaced 4$\times$4 grid of points, and destinations as random points around the origins (by up to 10\% of the image width in each direction).  These augmentations produce a 10-fold training data increase compared to the original dataset, while maintaining medical attributes \cite{perez2018data}.

\fakeparagraphnodot{CRACK} data set~\cite{ozgenel2017concrete} contains 40\,K images of 224$\times$224 pixels sliced from 500 full resolution images of 4032$\times$3024 pixels taken from the walls and floors of several concrete buildings. 
The images are taken approximately 1\,m away from the surfaces with the camera directly facing the target. The concrete surfaces have variation in terms of surface finishes (exposed, plastering and paint). The label is positive if an image contains a crack and negative otherwise. The labels are assigned by the material science experts. \figref{crack_pos_neg} shows positive and negative samples in this data set.

\subsection{Evaluation Metrics}
We adopt the following evaluation metrics: \aucroc, accuracy, False Negative Rate (FNR), and False Positive Rate (FPR) shortly summarized below. 

\fakeparagraphnodot{\aucroc} measure estimates the probability that a randomly chosen sample of a positive class has a smaller estimated probability of belonging to a negative class than a randomly chosen member of a negative class~\cite{steck2007hinge}.
: 
\begin{equation}
\text{\aucroc} = \dfrac{1}{n^+n^-} \sum_{i=1}^{n^+}\sum_{j=1}^{n^-}\mathbb{I}{(r_i^+ > r_j^-)},
\end{equation}
where $n^+, n^-$ are the number of positive and negatives samples and $\mathbb{I}$ is the indicator function. $r_i^+$ $\in {1, ..., n^+}$  denotes the rank of positive examples and $r_j^-$ $\in {1, ..., n^-}$  denotes the rank of negative examples.

%

We use the FNR measure to show that the proposed method indeed decreases the number of false negatives. We also report the FPR measure to understand the achieved trade-offs between the false positive and negative rates.
\aucroc measures the area underneath the entire ROC curve.  An ROC curve plots TPR vs. FPR at different classification thresholds.

The results in \cite{cortes2004auc} show that the expected value of \aucroc over all classifications is a monotonic function of accuracy. This also holds for imbalanced data. We report accuracy to ensure the overall compressed network performance stays high.

\subsection{Experimental Setup}
We enforce data imbalance in the CRACK data set by using 20\,K images in the negative class (no crack), and 4\,K images of the positive class, and use 70\,\%, 15\,\%, 15\,\% of samples for train, validation and test, respectively.

\fakeparagraph{Networks} For classification on the ISIC-2016 and CRACK data sets, we adopt AlexNet~\cite{krizhevsky2012imagenet} pre-trained on ImageNet~\cite{deng2009imagenet} with an adjusted number of fully-connected layers to contain 256, 8, and 2 neurons. This network has 2'471'842 parameters. Since our compression method uses iterative pruning based on the LT algorithm, we use these relatively shallow networks to keep the computation manageable on the available computational resources. The technical bottleneck here is the turn-off of the gradient in the backward pass in order to keep the pruned values set to zero. Given a stronger hardware infrastructure our method can be used on deeper networks such as VGG and ResNet. 

\fakeparagraph{Hyperparameters} 
To evaluate the performance of the proposed method, we follow our two-step design. First we focus on balancing the imbalanced data using the $\gamma$ parameter, then we use $\lambda$ multipliers for ranking loss to optimize \aucroc. 
Leveraging the results reported in~\cite{huang2016learning, cui2019class} to achieve data balancing by setting the loss function parameters according to the inverse class frequencies, we set $\beta$ close to $1$ in our tests. For ISIC and CRACK data sets $\gamma_c={727}/{173}=4.2$ and $\gamma_c={14K}/{2.8K}=5$ respectively. For simplicity, we use $\gamma_c=5$ in both cases. This corresponds to $\beta=0.99997$ for both data sets. We test $\lambda_c \in\{0, 1, 2, 10\}$, where $\lambda_c=0$ stands for the standard LT algorithm, and $\lambda_c=5$ performs best in all other scenarios with non-uniform weights for positive and negative classes. 

For each round of \algref{alg:LT}  in step 2 we train the network for $k=100$ iterations. With stronger hardware, it is possible to train the network longer to achieve potentially better results. Since our procedure trains an already pre-trained AlexNet on ImageNet dataset, the accuracy difference between $k=1000$ and $k=100$ iterations is less than 3\%.  By following the magnitude increase pruning strategy we prune $p=50$\% of the remaining weights in every round. This yields compressed networks with $|W_k| = $ 100\%, 50\%, 25\%, 12.5\%, 6.25\%. 3.12\%, and 1.57\% remaining weights. We use the stochastic gradient descent (SGD) with a momentum setting of $0.9$ for network training.

\fakeparagraph{Scenarios}
We test our method in three $\color{blue}blue$, $\color{black}black$, and $\color{ForestGreen}green$ scenarios, along with the $\color{red}LT$ algorithm as a baseline, and compare the obtained performance to three recent more popular benchmarks. These scenarios correspond to differently colored lines in the plots:
\begin{description}
\item[{\color{red}red}] Corresponds to the original LT algorithm with magnitude increase pruning criterion. The weights for both positive and negative classes are set to $\gamma_c=1$ and we use no ranking loss with $\lambda_c=0$. Thus, use the best performance of the classical LT algorithm as a benchmark. 
\item[{\color{blue}blue}] In this scenario, we test the effect of our ranking loss (class balancing) compared to the standard LT algorithm (red), without weighting cross entropy loss (data balancing). Therefore,  we have $\gamma_c=1$ and $\lambda_c=5$.
\item[{\color{black}black}] In this scenario, we test the effect of  weighting cross entropy loss (data balancing). Therefore, compared to the previous scenario (blue), we have $\gamma_c=5$ along with $\lambda_c=5$. We find that starting the first round with $\gamma_1=1$ helps the network to initially find the boundaries between the two classes without any specific focus
\item[{\color{ForestGreen}green}] In this scenario we test for weighting higher values for data balancing, so we set $\gamma_c=10$ everywhere, while $\lambda_c=5$.
\item[{\color{SeaGreen}LOBS}~\cite{dong2017learning}]  stands for the Layer-Wise Optimal Brain Surgeon algorithm. This pruning method determines the importance of neurons from the values in the corresponding Hessian matrix and prunes the least important ones in every layer.
\item[{\color{RedOrange}SNIP}]  is a single-shot network pruning algorithm~\cite{lee2018snip}. It prunes the filters based on a connection sensitivity criteria. Here we prune the network without re-training.
\item[{\color{Sepia}SNIP with training}]means a trained network is pruned by SNIP and then re-trained.
\item[{\color{Maroon}MobileNet}~\cite{howard2017mobilenets}] replaces the normal convolution by the depth-wise convolution followed by the point-wise convolution, which is called a depth-wise separable convolution. This reduces the total number of the floating point multiplication operations and hence significantly reduces the number of parameters.
\end{description}

\subsection{Results} 

\begin{figure}[t]
	\begin{subfigure}[b]{0.49\columnwidth}
	\includegraphics[width=\linewidth]{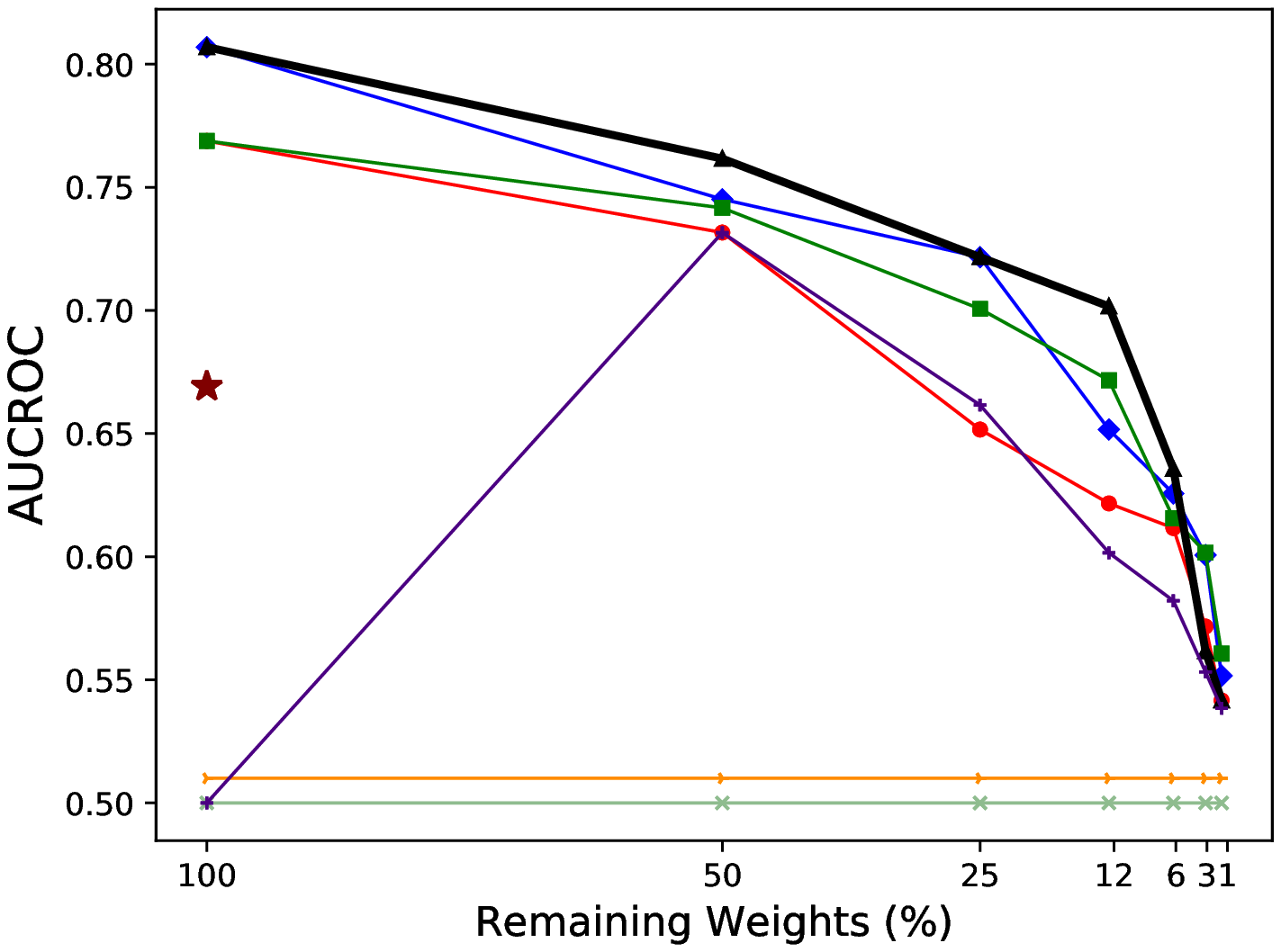}
	\vspace{-1.3\baselineskip}
	\caption{\aucroc}
	\label{isic:1}
\end{subfigure}
\hfill
	\begin{subfigure}[b]{0.49\columnwidth}
	\includegraphics[width=\linewidth]{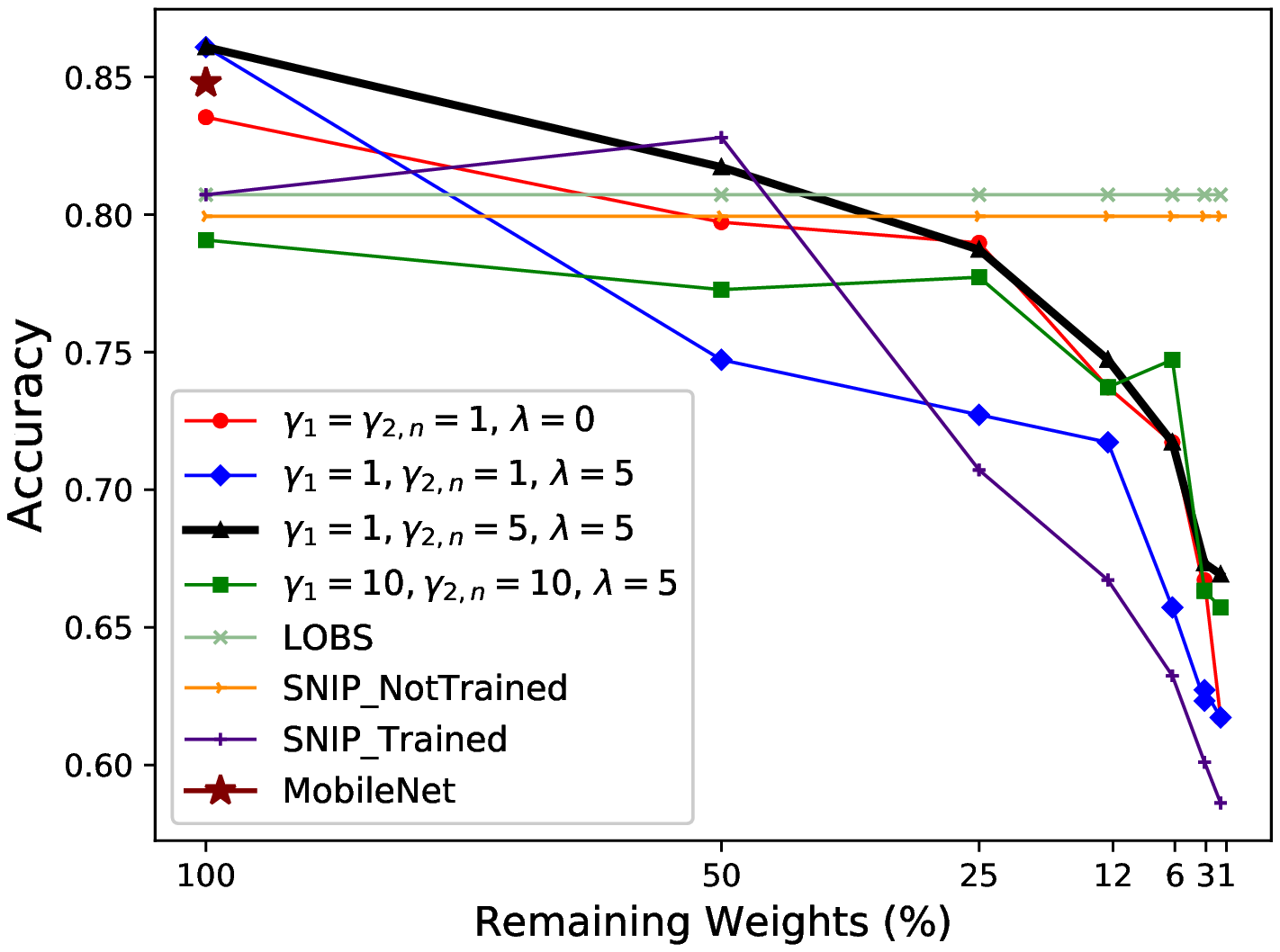}
	\vspace{-1.3\baselineskip}
	\caption{Accuracy}
	\label{isic:2}
\end{subfigure}
\begin{subfigure}[b]{0.49\columnwidth}
	\includegraphics[width=\linewidth]{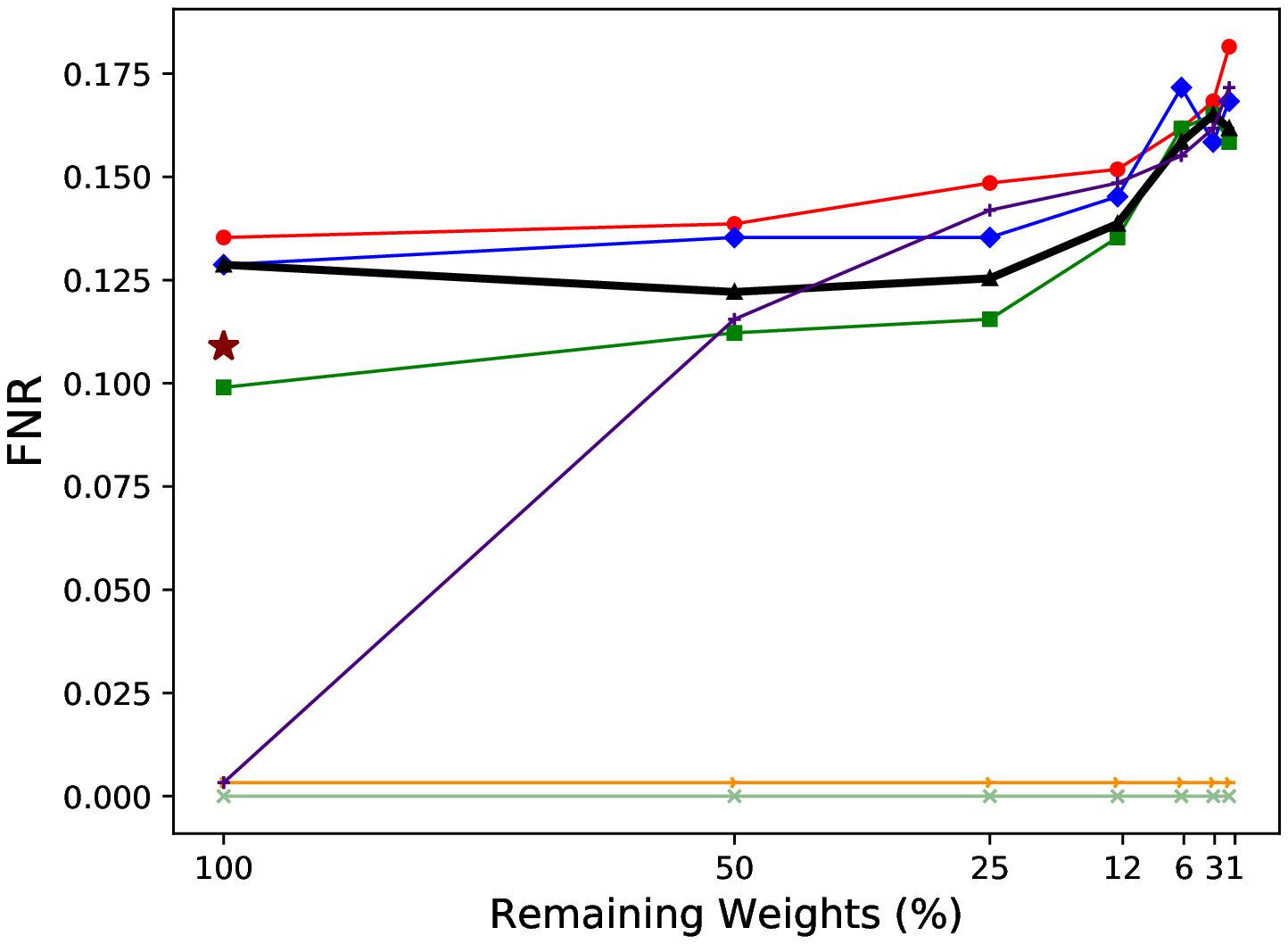}
	\vspace{-1.3\baselineskip}
	\caption{FNR}
	\label{isic:3}
\end{subfigure}
\begin{subfigure}[b]{0.49\columnwidth}
	\includegraphics[width=\linewidth]{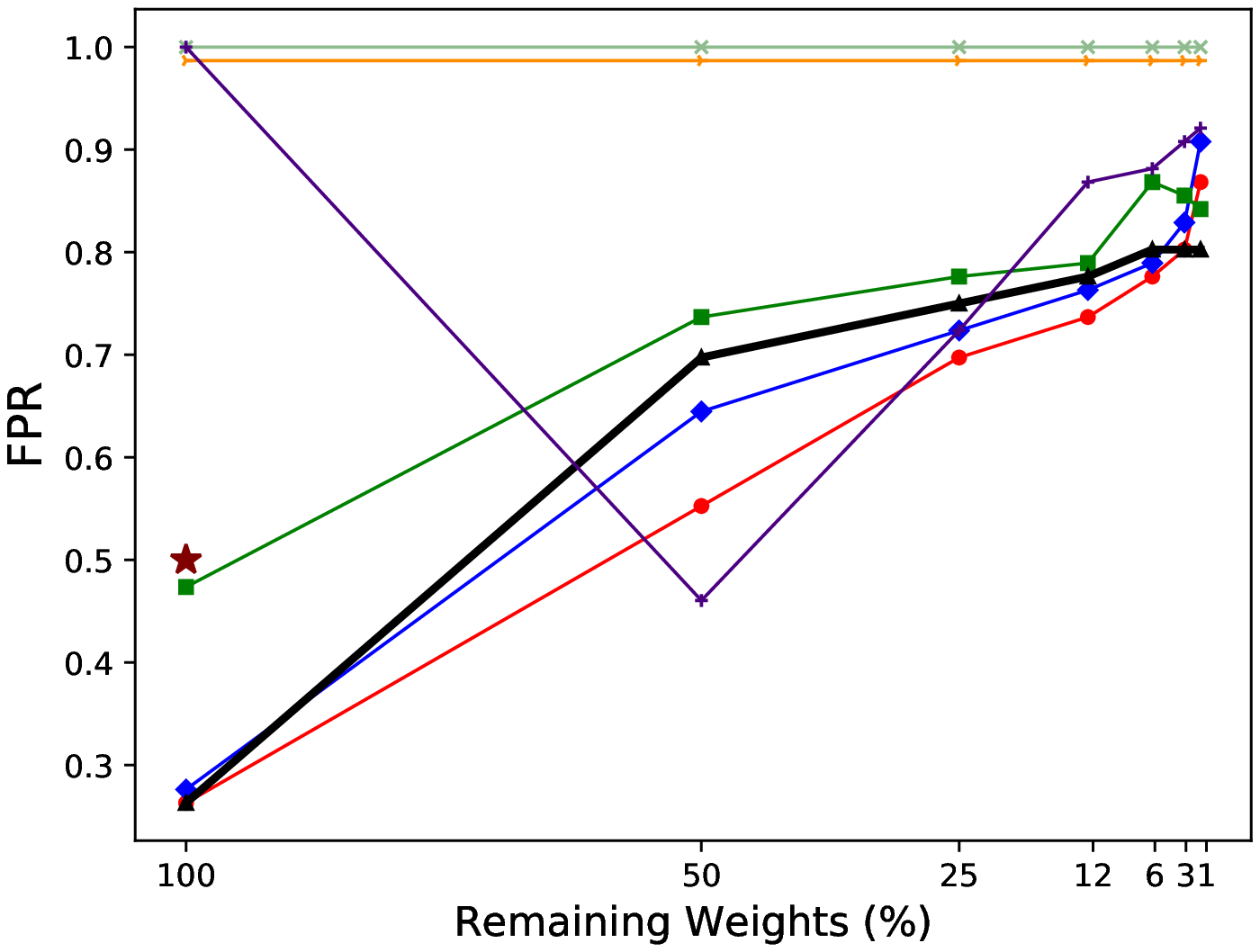}
    	\vspace{-1.3\baselineskip}
	\caption{FPR}
    	\label{isic:4}
\end{subfigure}
\caption{Evaluation results on ISIC-2016. Our method (black line) outperforms the LT algorithm (red line) in \aucroc, accuracy, FNR, and FPR for up to 1\% of remaining weights in the pruned network. Our method also outperforms LOBS, SNIP, and MobileNet in \aucroc and FPR.}
\label{fig:isic_eval}
\end{figure}

\begin{figure}[t]
\begin{subfigure}[b]{0.49\columnwidth}
	\includegraphics[width=\linewidth]{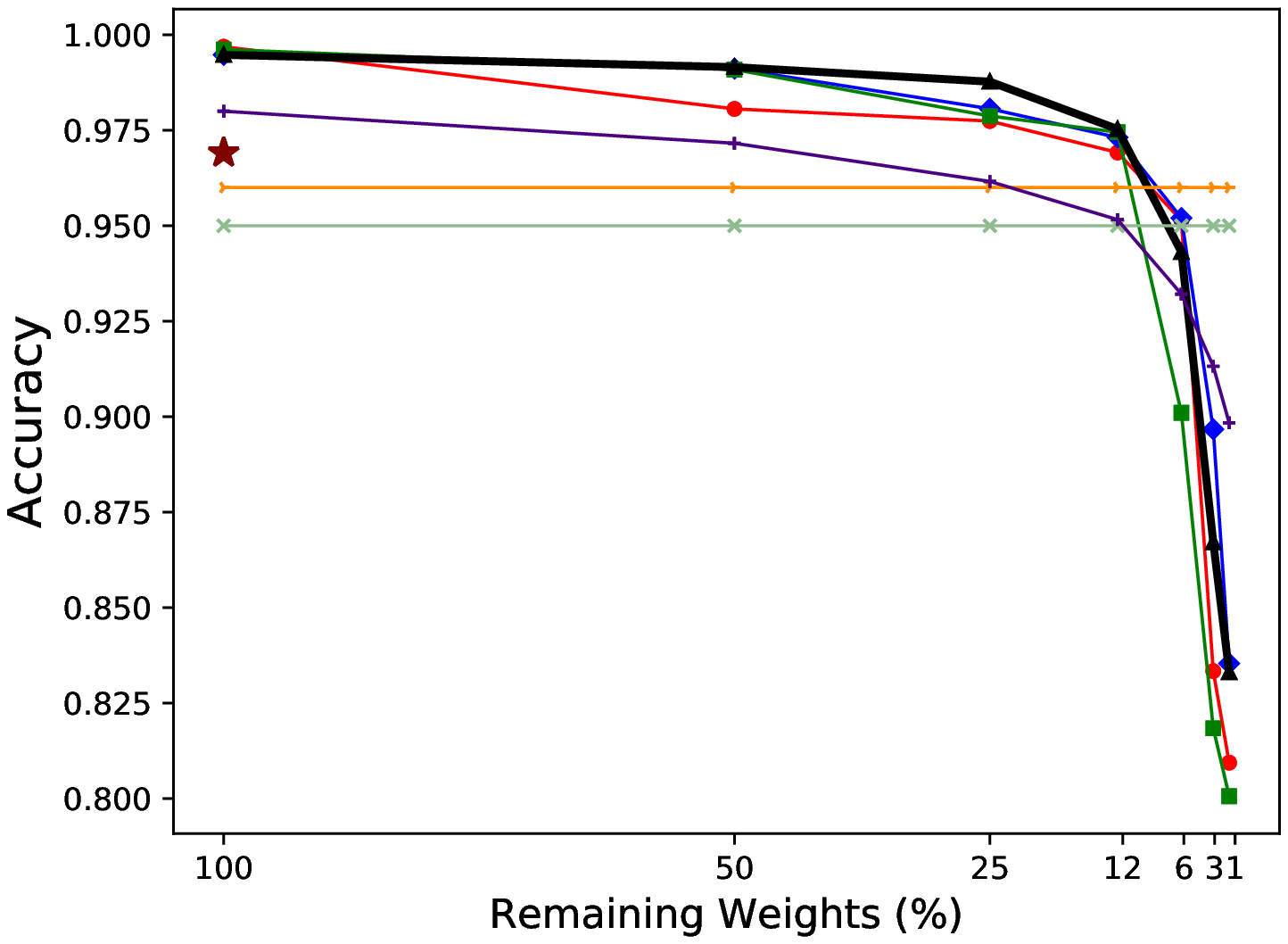}
    	\vspace{-1.3\baselineskip}
	\caption{\aucroc}
	\label{crack:2}
\end{subfigure}
\begin{subfigure}[b]{0.49\columnwidth}
	\includegraphics[width=\linewidth]{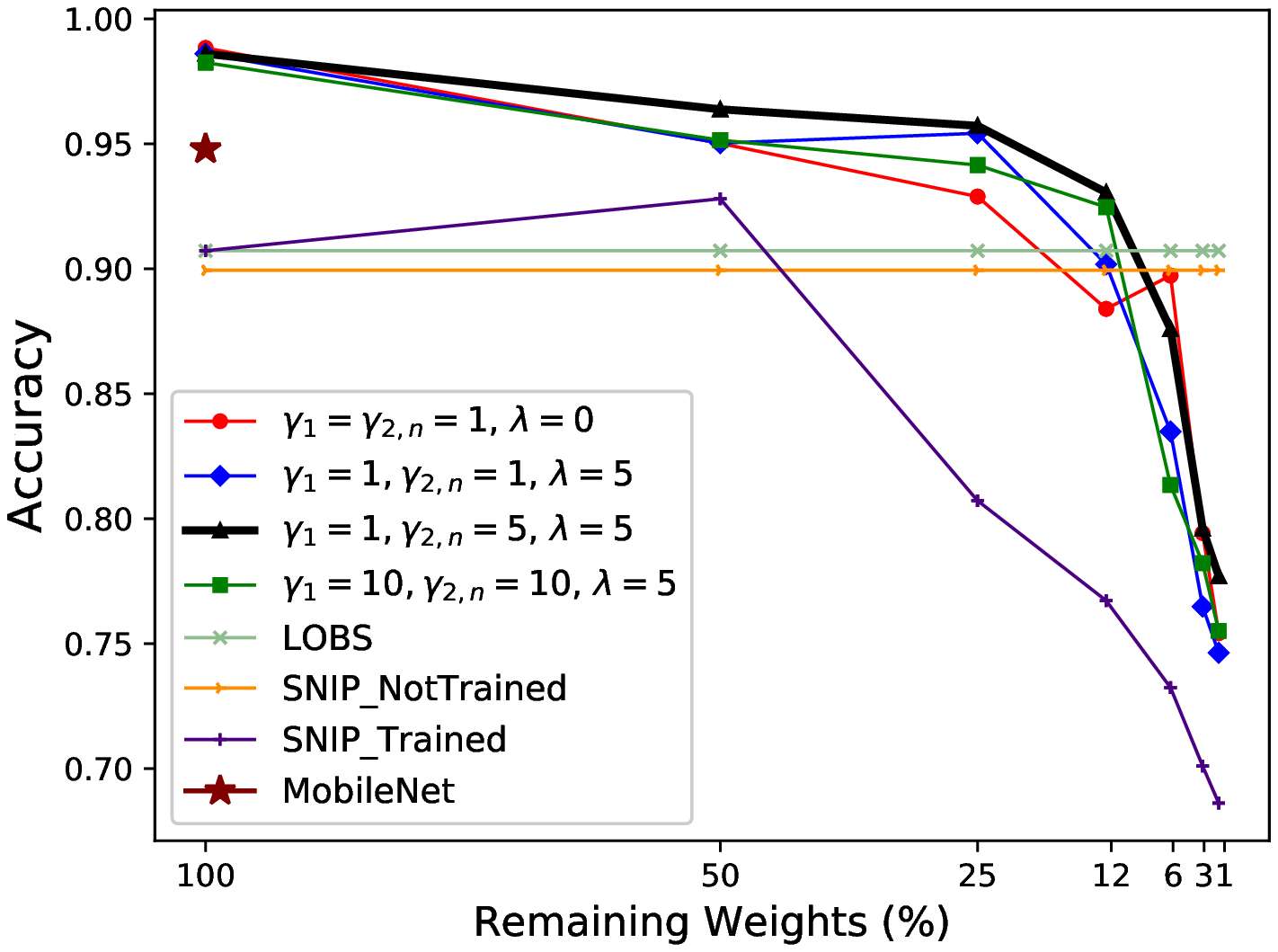}
	\vspace{-1.3\baselineskip}
	\caption{Accuracy}
	\label{crack:4}
\end{subfigure}
\begin{subfigure}[b]{0.49\columnwidth}
	\includegraphics[width=\linewidth]{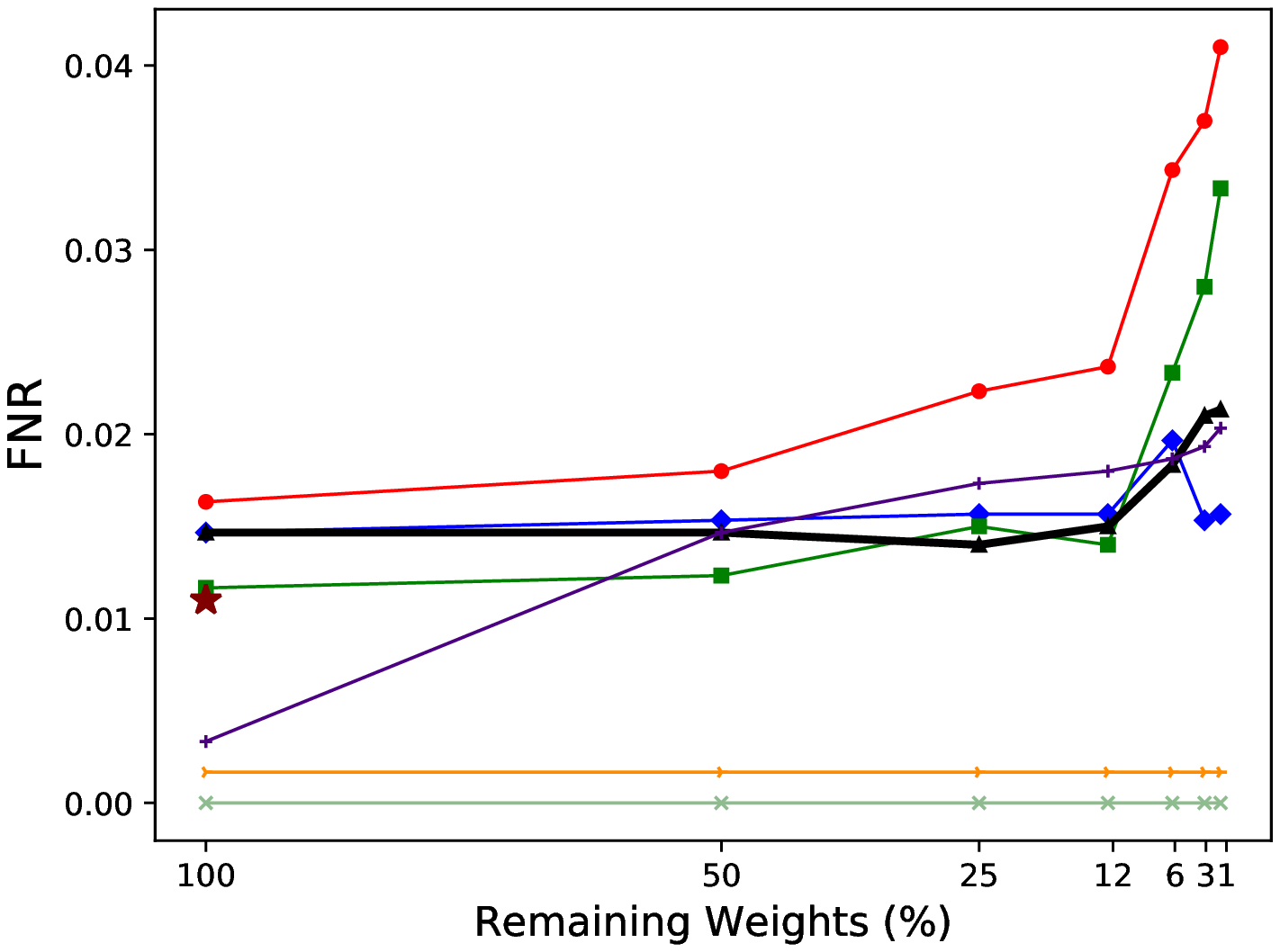}
	\vspace{-1.3\baselineskip}
	\caption{FNR}
	\label{crack:5}	
\end{subfigure}
\begin{subfigure}[b]{0.49\columnwidth}
	\includegraphics[width=\linewidth]{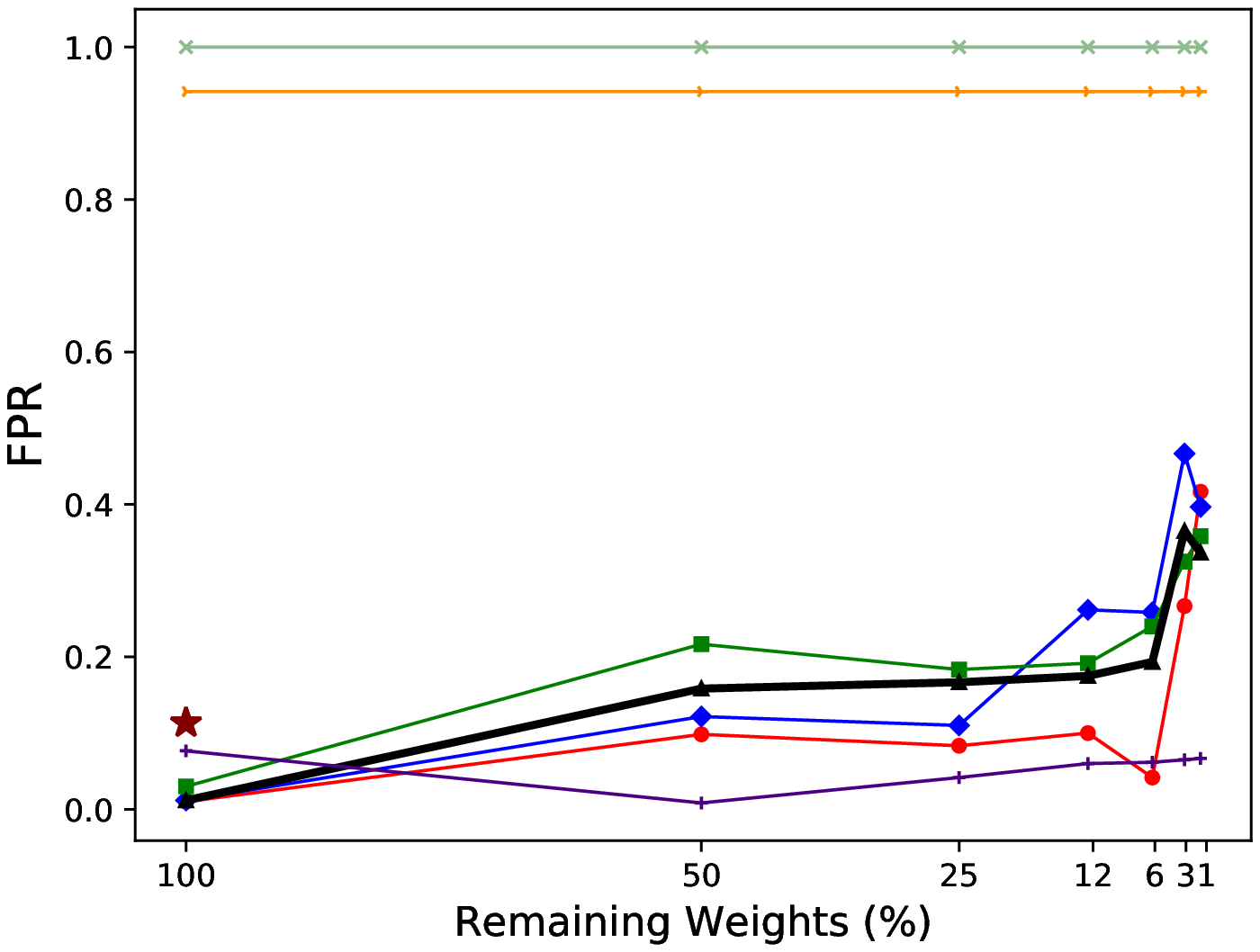}
	\vspace{-1.3\baselineskip}
	\caption{FPR}
	\label{crack:6}
\end{subfigure}
\caption{Evaluation results on the CRACK data set. Our method (black line) outperforms the LT algorithm (red line) in \aucroc, accuracy, FNR, and FPR for up to 12\% of the remaining weights. We also outperform LOBS, SNIP and MobileNet in \aucroc and accuracy.}
\label{fig:crack_eval}
\end{figure}
\figref{fig:isic_eval} and \figref{fig:crack_eval} show the evaluation results for ISIC and CRACK data sets. We compare our results to the best performance obtained with the classical LT algorithm with the magnitude increase pruning criterion along with three other recent benchmarks. Our goal is to prune the network while maximizing the \aucroc measure and the classification accuracy, yet keep the number of false negatives for a specific class as low as possible.

As can be seen in \figref{fig:isic_eval}(\figsubref{isic:1}) the best \aucroc for ISIC (black line) is achieved when $\lambda_c=5$, $\gamma_c=1$ for the first training epoch ($\gamma_1=1$) and $\gamma_c=5$ for later epochs ($\gamma_{2,n}=5$) , where $\gamma_c=5$ gives the inverse class frequency for this data set.
The setting $\gamma_1=1$ highlights the importance of learning the class boundaries in the first iteration by using balanced training. The focus on the positive class therefore starts from the second iteration where we use $\gamma_{2,n}=5$ for the desired positive class. This setting not only outperforms the standard lottery ticket in \aucroc, but also recent popular benchmarks, \ie LOBS, SNIP, and MobileNet.

On the one hand, the presented results in \cite{cortes2004auc} show that the expected value of the \aucroc over all classifications is a monotonic function of accuracy, when we have an imbalanced dataset. On the other hand, authors in \cite{cui2019class} argue that the data-balancing term $\gamma$ improves the performance of the cross-entropy loss in terms of accuracy. Our results in \figref{fig:isic_eval}(\figsubref{isic:2}) confirm these findings: the accuracy for the best settings of our method (black line) is consistently better than the accuracy of the LT algorithm (red line). Although the accuracy for LOBS, and SNIP remains high through compression, this is due to the fact that they only correctly classify all the samples in the positive class (see high FPR in \figref{fig:isic_eval}(\figsubref{isic:4})).

\figref{fig:isic_eval}(\figsubref{isic:3}) shows that adding ranking loss to the standard LT algorithm as a proxy for class-balance improves the FNR (blue line). However, weighting the cross entropy for data-balance improves the FNR even stronger (black line). The best FNR is achieved when we use a higher weight for the positive class by setting $\gamma_c=10$ in all pruning rounds (green line). The Standard LT algorithm and the green method yield FNRs of 0.13 and 0.09, which shows 35\% improvement over LT. Comparing \figref{fig:isic_eval}(\figsubref{isic:4}) with \figref{fig:isic_eval}(\figsubref{isic:3}) shows a trade-off between FPR and FNR. As we can see here, we have the lowest FPR, when we have no class-balance and no data-balance. The FPR for LOBS and SNIP without training (after pruning) are also very high (close to 1), meaning they incorrectly classify all the nagative samples as positive.

Our best setting (black line) in the first iteration (without compression) beats the best \aucroc and accuracy for the ISIC challenge: Our \aucroc and accuracy are $0.8069$ and $0.8608$ respectively, which is superior to \aucroc=$0.804$ and accuracy=$0.855$ reported by the authors in \cite{yu2016automated}. Their proposed very deep architecture is highly dependent on the segmentation results, whereas our method is an end-to-end algorithm. \\
For the CRACK data set, \figref{fig:crack_eval}(\figsubref{crack:2}) shows the results for \aucroc. The  best \aucroc is achieved when using the following set of parameters: $\gamma_1=1$, $\gamma_{2,n}=5$, $\lambda_c=5$ (black line), where $\gamma_c=5$ gives the inverse class frequency for this data set. Similar to ISIC, our method outperforms the standard LT, and the popular benchmarks LOBS, SNIP, and MobileNet in terms of \aucroc.
As can be seen in \figref{fig:crack_eval}(\figsubref{crack:4}) the accuracy for our method in the best setting (black line), which is not only better than the best accuracy achieved by the LT method (red line), but also LOBS, SNIP and MibileNet. Apart from LOBS and SNIP without training (after pruning) which have FPR close to 1, the best achieved FNR in \figref{fig:crack_eval}(\figsubref{crack:5}) is when we set $\gamma_c=10$ for all pruning rounds (green line). FNR for the LT algorithm and the green line yield FNR of 0.017 and 0.011, which shows 35\% improvement. However, due to the natural trade-off between FNR and FPR, giving more weight to the positive class leads to a higher FPR.

\section{Related Work}
\label{sec:related}
Deep networks are known to be highly redundant. This motivated many researchers to seek for network compression techniques and efficient subnetworks. 
This section summarizes recent efforts. 

\fakeparagraphnodot{Quantization and binarization} rely on weights with discrete values. 
\cite{soudry2014expectation} propose an algorithm, which approximates the posterior of the neural network weights, yet the weights can be restricted to have discrete or binary values. 
\cite{rastegari2016xnor} apply approximations to standard CNNs. Their Binary-Weight-Network approximates the filters with binary values and reduces the size of example networks by the factor of 32x. 

\fakeparagraphnodot{Decomposition and factorization} explore low-rank basis of filters to reduce model size.
\cite{jaderberg2014speeding} represent the learnt full-rank bank of a CNN as a combination of rank-1 filters leading to a 4.5x speed-up. 
More recent methods \cite{mehta2019espnetv2} rely on depth-wise and point-wise separable convolutions to reduce computational complexity. Depth-wise convolution performs light-weight filtering by applying a single convolutional kernel per input channel. Point-wise convolution expands the feature map along channels by learning linear combinations of the input channels. 

\fakeparagraphnodot{Pruning} covers a set of methods which reduce the model size by removing network connections. 
These methods date back to the optimal brain damage~\cite{lecun1990optimal}, where the authors suggest to prune the weights based on the Hessians of the loss function. 
\cite{li2016pruning} propose to prune the channels in CNNs based on their corresponding filter weight norm, while \cite{hu2016network} uses the average percentage of zeros in the output to prune unimportant channels. 
The LT hypothesis~\cite{frankle2018lottery} proposes iterative pruning to remove the weights with small magnitude. The methods often yields efficient sparse sub-networks.

\fakeparagraphnodot{Knowledge distillation} covers methods which transfer knowledge from a larger \emph{teacher} to a smaller \emph{student}.
\cite{belagiannis2018adversarial} exploit adversarial setting to train a student network. The discriminator tries to distinguish between the student and the teacher. They use the L2 loss to force the student to mimic the output of the teacher.
\cite{aguinaldo2019compressing} apply knowledge distillation to GANs to produce a compressed generator without neither loss of quality nor generalization. They hypothesized that there exists a fundamental compression limit of GANs similar to Shannon's compression theory.

Our network compression method combines network pruning with efficient network design. Unlike other compression methods, which try to minimize the overall error rate, our method additionally optimizes \aucroc while focusing on a desired class which appears to be useful in a number of real applications in the IoT domain. 


\bibliographystyle{IEEEtran}

\end{document}